\documentclass[reprint, cha]{revtex4-1}
\usepackage{amssymb}
\usepackage{graphicx}
\usepackage{amsmath}
\usepackage{amsfonts}
\usepackage{amssymb}
\usepackage{amsthm}
\usepackage{verbatim}
\usepackage{enumerate}
\usepackage{natbib}
\usepackage{mdwlist} 
\usepackage{subfigure}
\usepackage{color}
\usepackage[utf8]{inputenc}

\usepackage{algpseudocode}
\usepackage{algorithmicx}
\usepackage{lineno}

\begin{document}

\title{Texture analysis using deterministic partially self-avoiding walk with thresholds}

\author{Lucas Correia Ribas}
 	     \email{lucasribas@usp.br}
\affiliation{Institute of Mathematics and Computer Science, University of S\~{a}o Paulo (USP), Avenida Trabalhador s\~{a}o-carlense, 400 13566-590 S\~{a}o Carlos, S\~{a}o Paulo, Brazil} 
\affiliation{Scientific Computing Group, S\~ao Carlos Institute of Physics, University of S\~{a}o Paulo (USP),  PO box 369 13560-970 S\~{a}o Carlos, S\~{a}o Paulo, Brazil - www.scg.ifsc.usp.br}

\author{Wesley Nunes Gon{\c{c}}alves}
\email{wesley.goncalves@ufms.br}
\affiliation{Federal University of Mato Grosso do Sul, Rua Itibir\'e Vieira, s/n, 79907-414, Ponta Por\~a, MS, Brazil }

\author{Odemir Martinez Bruno}
              \email{bruno@ifsc.usp.br}
\affiliation{Scientific Computing Group, S\~ao Carlos Institute of Physics, University of S\~{a}o Paulo (USP),  PO box 369 13560-970 S\~{a}o Carlos, S\~{a}o Paulo, Brazil - www.scg.ifsc.usp.br}

\date{\today}

\begin{abstract}

	In this paper, we propose a new texture analysis method using the deterministic partially self-avoiding walk performed on maps modified with thresholds. In this method, two pixels of the map are neighbors if the Euclidean distance between them is less than $\sqrt{2}$ and the weight (difference between its intensities) is less than a given threshold.
	The maps obtained by using different thresholds highlight several properties of the image that are extracted by the deterministic walk.
	To compose the feature vector, deterministic walks are performed with different thresholds and its statistics are concatenated.
	Thus, this approach can be considered as a multi-scale analysis.
	We validate our method on the Brodatz database, which is very well known public image database and widely used by texture analysis methods.
	Experimental results indicate that the proposed method presents a good texture discrimination, overcoming traditional texture methods.
\end{abstract}


\maketitle

	\section{Introduction}
	
	Texture analysis is an important research field in computer vision and image processing with applications in various areas, such as analysis of medical images \cite{Ramamoorthy2015747}, recognition of plant leaves \cite{BackesB10a,BackesCB09}, quality control of food \cite{Chen2014285}, remote sensing \cite{zhu1998study}, geological images \cite{Heidelbach200091}, microscope images
	\cite{anguiano1999surface}, etc. Texture is a visual attribute widely used to describe patterns in images by a wide variety of descriptors proposed along the years \cite{Goncalves201211818,backes2010texture}.
	Although its understanding and perception are natural for humans, its formal definition proved to be a hard task \cite{Goncalves20132953,Ribas2015}.
	Generally, the definitions describe the textures as repetitive and regular patterns, however, they are much more complex. Textures can present patterns combined in different scales, absence of patterns (e.g., noises), or complex patterns, such as most part of the natural textures. 
	
	A range of methods for texture analysis has been proposed over the years. These methods can be divided into five categories: statistical methods, structural methods, spectral methods, model-based methods and agent-based methods.
	Statistical methods describe textures in images by means of first and second order statistical measures that are related to gray levels of the pixels.
	One of the most popular methods is the Haralick descriptors \cite{haralick}, which is based on gray level co-occurrence matrix (GLCM) to extract statistical measures, such as, contrast, correlation, energy, entropy, etc. 
	Examples of methods of this category include Laplacian of Gaussian \cite{voorhees87} and Lazebnik descriptor \cite{lazebnik}.
	Spectral methods convert the image in a new image based on properties of spatial frequencies of the intensity of the pixels \cite{bharati2004image}. A classic method is the Fourier descriptor \cite{azencott1997texture}, that applies the Fourier transform over the image and extract characteristics in the frequency domain. Another methods are Gabor filters \cite{Bovik199055} and Wavelet transform \cite{Unser19951549}.
	Model-based methods are based on the construction of image models that use the parameters estimated to describe the texture. Generally, the models used are stochastic and fractals. The main methods are based on Gaussian Markov random fields \cite{Cross198325}, fractal dimension of Bouligand-Minkowski \cite{Backes20135870,BackesCB12,Ribas2015}, multi-fractal \cite{xu2009viewpoint} and multi-scale fractal dimension \cite{FlorindoB13}.
	More recently, agent-based methods have been proposed for texture analysis.
	These methods use an individual guided by a deterministic or stochastic rule, which performs walks in the image and uses information of the walk as feature vector. 
	The main advantage of this category compared to others literature methods is its capacity of describing micro-textures in an effective way, such as micro-textures in leaf texture images that have small texture variations between species \cite{Goncalves20132953}.
	Methods of this category include deterministic tourist walk \cite{backes2010texture,backesTourist,Goncalves201211818,Goncalves20132953} and random walk \cite{Goncalves201651}.
	
	In earlier papers, the deterministic partially self-avoiding walks (or deterministic tourist walk - DPSW) have been used to describe macro and micro textures \cite{Goncalves201211818,Goncalves20132953,Backes20111684,backes2010texture,backesTourist}. A partially self-avoiding walk can be understood as a tourist that wants to visit cities distributed in a d-dimensional map according to a deterministic rule and a given memory. For texture analysis, the image is transformed into the map of the tourist, as follows: each pixel is considered a city, the neighborhood of a pixel is the 8-connected and the "distance" between two pixels is given by the module of the difference of their gray levels (weight). Thus, a deterministic partially self-avoiding walk is initiated at each pixel of the image and follows the deterministic rule: goes to the pixel that minimizes the weight and
	that has not been visited in last $\mu$ steps \cite{Goncalves20132953,backes2010texture}.
	Each walk produces a trajectory that has an initial transient of size $\tau$ and ends with an attractor of size $\rho$. The transient is an initial sequence of pixels until finding an attractor. The attractor is a cycle of pixels from where the tourist cannot escape. For texture characterization, statistical measures are obtained from a joint distribution that computes the frequency of trajectories with a certain size of transient and period of attractor.

	Unlike from earlier works, in this paper, we propose a new approach to perform DPSW on a map modified by thresholds over the weights of the neighbors. Thus, in this work, we changed the function of neighborhood so that the tourist walk does not consider in the next step the pixels whose weight is less than a given threshold.	
	The application of thresholds on the neighbors changes the rule of movement, which produces new trajectories of deterministic walk and explores new properties of the image.
	Thus, the use of a set of thresholds can be considered a multi-scale approach that highlights different scales and information of the image.
	Experiments using textures from the Brodatz database indicate that our method adds new information for texture analysis and increases significantly the classification compared to the traditional DPSW method. In addition, the proposed method presents better results than the other traditional texture methods.
	
	This paper is organized as follows.
	Section \ref{sec:dpsw} presents a review about the deterministic partially self-avoiding walk. 
	In Section \ref{sec:proposed}, we describe the proposed method.
	Experimental results using images extracted from the Brodatz database are presented in Section \ref{sec:exp}. Section \ref{sec:conc} concludes the paper and discusses future works.

	\section{Deterministic partially self-avoiding walk} \label{sec:dpsw}
	
	Recently, the DPSW emerged as a very promising approach in computer vision, with interesting results in static textures analysis (e.g., \cite{backes2010texture,Backes20111684,Goncalves201211818,Goncalves20132953}) , dynamic textures analysis (e.g. \cite{Goncalves20131163,Goncalves20134283}) and recognition of plant leaves  (e.g., \cite{backesTourist}).
	The first work proposed for image analysis is described in Ref. \cite{campiteliTourist}. This method uses the transient and attractor sizes for texture representation. Then, Backes et. al. \cite{backesTourist} proposed a new variation that combines two rules of movement and a new measure extracted from the joint distribution. Besides that, the developed method was applied in the recognition of plant species through their image.

	In Backes et al. \cite{Backes:2010:TAB:1840008.1840249}, another variation was proposed to solve the tie problem in the walk according to a new rule of movement based on the neighborhood contrast. The experimental results show interesting performance in rotation images classification compared to other methods of the literature. In Ref. \cite{Goncalves20132953}, a new approach combines the fractal dimension and DPSW for texture representation. This study uses fractal descriptors to analyze the incidence, distribution and spatial arrangement of the attractors.
	Similarly, an extension of the DPSW was proposed to be combined with the graph theory \cite{Goncalves201211818,Backes20111684}. In dynamic texture analysis, two approaches based on DPSW were proposed for recognition, clustering and segmentation with applications in problems of nanotechnology and videos of traffic \cite{Goncalves20134283,Goncalves20131163}.
	
	\subsection{Definition}
	
	DPSW can be understood as a tourist which wants to visit $N$ cities distributed on a map of $d$ dimensions \cite{backesTourist}. The tourist starts the walk in a predefined city with a rule of movement and memory, which drives it to next city: go to the nearest city that has not been visited in the last $\mu$ steps \cite{backes2010texture,Backes20111684}. The memory $\mu-1$ is the window in which the tourist performs a deterministic partially self-avoiding walk, i.e., the tourist considers a city attractive again, only after visiting $\mu$ cities. The rule of movement is based in the neighborhood and a small memory, that although simple can produce DPSW of great complexity \cite{backes2010texture}.

	The DPSW can be separated into two parts: transient and attractor. The transient is the initial part of the walk that has size equal to $\tau$. The attractor is the final part, composed by a cycle of cities of size $\rho \geq \mu+1$ in that the tourist gets stuck. This is the attractor is a cycle of cities that always will be visited by the tourist. Thus, the transient is the route of the tourist until finding an attractor. In some cases, it can occur of the tourist do not find an attractor, therefore, the trajectory of the tourist is only composed by the transient. In the case of existing various neighboring cities which satisfy the rule of movement, it is considered the first city that satisfies the rule in clockwise \cite{backes2010texture}. 
	
	For application of the DPSW on images, consider a bidimensional image composed by a finite set of pixels $\varphi$ in which each pixel has a location $p_i=(x_i,y_i)$ and a gray level $I(p_i)\in[0,255]$. Each pixel $p_i$ is considered a city and the neighborhood $\eta(p_i)$ of the pixel $p_i$ is the pixels whose the Euclidian distance is smaller or equal than $\sqrt{2}$, according to Equation \ref{eq:dist}. This is, the neighborhood used by the traditional methods is the 8-connected neighborhood.
	
	\begin{equation}
	\begin{aligned}
	\eta(p_i) = \{p_j \mid d(p_i,p_j) \leq \sqrt{2}\}, \\
	d(p_i,p_j)=\sqrt{(x_i-x_j)^2 + (y_i-y_j)^2}
	\label{eq:dist}
	\end{aligned}
	\end{equation} 
	
	Once two pixels $p_i$ and $p_j$ are considered geometric neighbors, the real "distance" between them is given by a weight $w_{p_i,p_j}$. This weight is defined as the module of the difference of their gray levels $w_{p_ip_j}=\mid I(p_i)-I(p_j) \mid$ \cite{backes2010texture}.
	
	Given the above definitions, the methods for texture analysis based on DPSW considers independent tourists walking through image pixels. Thus, the image is considered the map to perform the tourist walks.
	Given that a tourist is in pixel $p_i$ the rule of movement is defined as go to the pixel $p_j \in \eta(p_i)$ that minimizes the weight $w_{p_i p_j}$ and that has not been visited in the last $\mu$ steps. This rule is called $min$, she drives the tourist for similar pixels and explores homogeneous regions on the image. In addition, it is considered the rule of movement $max$. This rule moves the tourist to the pixel that maximizes the weight $w_{p_i p_j}$. In this rule, the tourist is guided for different pixels, exploring heterogeneous regions on the image (e.g., high-contrast regions and edges).
	
	%
	%
	
	The trajectory of the tourist is directly associated with the pixel where it is started, therefore, each pixel of the image is considered as an initial point to perform a walk.
	Thus, in an image of $N$ pixels, it is obtained a set of $N$ trajectories produced by the walks. To measure statistically the set of trajectories performed on an image, it is used a joint distribution $S_{\mu,r}(\tau,\rho)$. This distribution computes the frequency of trajectories that obtained transient size $\tau$ and period of attractor $\rho$ (Equation \ref{eq:dist_walk}) \cite{Goncalves20131163,Goncalves201211818,backes2010texture}.

	\begin{equation}\label{eq:dist_walk}
	S_{\mu,r}(\tau,\rho) = \frac{1}{N} \sum_{i=1}^{N} \left\{\begin{matrix}
	1,& \text{if } \tau_i = \tau \text{ e } \rho_i = \rho \\ 
	0,& \text{otherwise}
	\end{matrix}\right.
	\end{equation}
	where $i$ is the pixel that starts a walk and $r$ is the rule of movement.

	\section{Proposed Method}
	\label{sec:proposed}
	In this section, we describe the proposed method.
	The main difference of the proposed method for other works based on deterministic walkers is the map to perform the DPSW. As can be seen in Section \ref{sec:dpsw}, the map is composed by the image pixels and the neighborhood of a pixel is given by the function $\eta(p_i)$. This map is the same for the two rules of movement. In the proposed method, a threshold of cutting is applied over the neighborhood of the pixels to obtain a map that reveals different texture properties. Thus, different maps are considered according to each rule of movement. For the rule of movement $r=min$, the neighborhood of a pixel $p_i$ is composed of pixels whose the weight $w_{p_i p_j}$ is higher than $t_{min}$, according to
	
	\begin{equation}\label{eq:limiar_min}
	\eta^{min}_t(p_i)=   \{p_j \mid d(p_i,p_j) \leq \sqrt{2}, w_{p_i p_j} \geq t_{min} \}.
	\end{equation}
	
	Similarly, a neighborhood is also defined to the rule of movement $r=max$. This neighborhood $\eta^{max}_t(p_i)$ consists of pixels whose the weight $w_{p_i p_j}$ is below than $t_{max}$, according to
	
	\begin{equation}\label{eq:limiar_max}
	\eta^{max}_t(p_i)=   \{p_j \mid d(p_i,p_j) \leq \sqrt{2}, w_{p_i p_j} \leq t_{max} \}.
	\end{equation}
	
	The thresholds $(t_{min}=k)$ and $(t_{max}=255-k)$ are in function of a given $k$ value. As the value of $k$ increases, the threshold $t_{min}$ increases, on the other hand, the threshold $t_{max}$ decreases. Thus, it is possible to analyze the two rules of movement together. 
	The application of the thresholds over the neighborhood can be considered as a multi-scale texture analysis. For each threshold $k$, the original map (i.e., 8-connected neighborhood) is transformed into a $k$-scaled map. Compared to the original map, a $k$-scaled map presents different properties and reveals the information related to its scale. 
	Let us consider the rule of movement $r=min$.  For small values of $t_{min}$ the tourist will provide trajectories that models the homogeneous details of the image.
	As the value of $t_{min}$ increases, the tourist walk models global information, such as edges of the image. 
	For the rule of movement $r=max$ the idea is similar. 
	However, in this rule as the value of $t_{max}$ decreases, local information are highlighted.
	Thus, the proposed method aims at combining features from some $k$-scaled maps to achieve a multi-scale texture analysis.

\begin{figure*}[!htbp]
		\centering
		\includegraphics[width=1\textwidth]{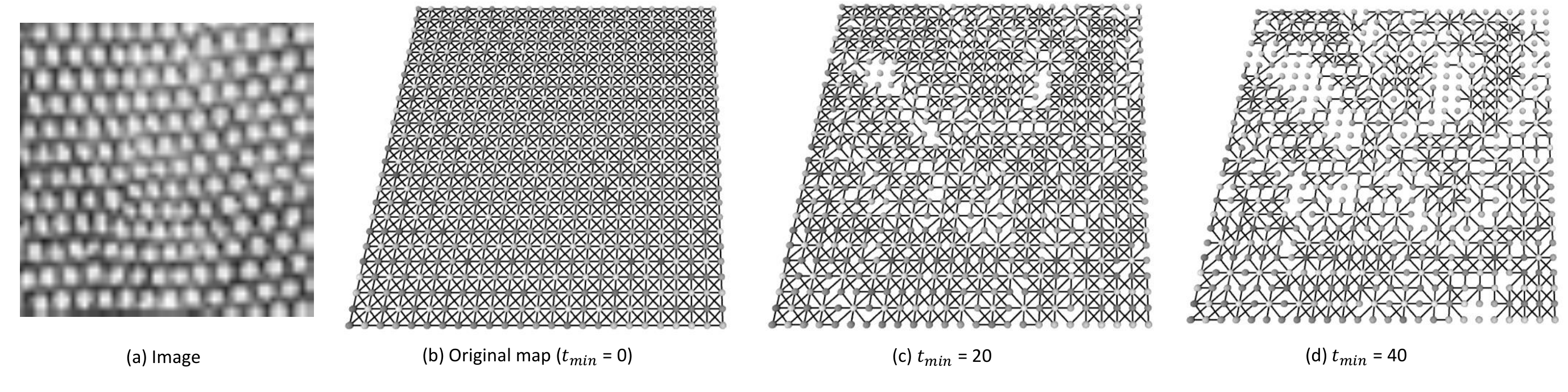}		
		\caption{Examples of maps obtained from an image using the neighborhood $\eta^{min}_t(p_i)$.}
		\label{fig:example}
	\end{figure*}

	Figure \ref{fig:example} shows examples of maps to perform the DPSW using the neighborhood $\eta^{min}_t(p_i)$.  In this figure, the pixels are represented by points with its grayscale and two pixels are neighbors if it has a connection between them.  Figure \ref{fig:example}(b) shows the original map used by the traditional methods of DPSW for texture analysis. Figures \ref{fig:example}(c) and \ref{fig:example}(d) show maps with neighborhoods built with thresholds $t_{min}=20$ and $t_{min}=40$, respectively. Notice that to increase the threshold $t_{min}$, homogeneous regions are disconnected taking the tourist to explore different regions and properties of the image.

	\subsection{Feature Vector}
	
	To use the DPSW for texture analysis and classification, an approach extensively studied in previous works is the feature extraction from the transient time and attractor period joint distribution \cite{Goncalves201211818,Backes20111684,backes2010texture}.	The histogram $h_{\mu,r}^k(l)$ proved to achieve better results. This histogram $h_{\mu,r}^k(l)$ summarize the number of trajectories that have a size equal to ($l=\tau+\rho$) in the joint distribution calculated for a memory $\mu$ and a rule of movement $r$ (Equation \ref{eq:hist}).
	
	\begin{equation}\label{eq:hist}
	h_{\mu,r}^k(l)= \sum_{b=0}^{l-1} S_{\mu,r}^k(b,l-b)
	\end{equation}
	
	To accomplish the task of classification, an interesting strategy is to build a feature vector $\nu_{\mu}^r$ with a total of $n$ descriptors selected from the histogram $h_{\mu}^r(l)$. Once there are no attractors of size smaller than ($\mu+1$), the first position of the histogram selected is ($\mu+1$) \cite{backes2010texture,Goncalves20134283}. Here, we use a total of $n=4$ histogram descriptors. Thus, the feature vector $\nu_{\mu}^r$ is calculated for a specific $\mu$ value, according to
	
	\begin{equation}\label{eq:vector}
	\nu_{\mu,r}^k = [h_{\mu,r}^k(\mu+1), h_{\mu,r}^k(\mu+2),...,h_{\mu,r}^k(\mu+n) ].
	\end{equation}
	
	To model different properties, we also propose a feature vector $ \varphi^k_r$ considering different $\mu$ values. The use of different memory sizes and rules of movement to characterize texture patterns presents a great potential for image classification \cite{backes2010texture}. The feature vector $\varphi^k_r$ is given by the concatenation
	
	\begin{equation}\label{eq:vector2}
	\varphi^k_r= [\nu_{\mu_1,r}^k , \nu_{\mu_2,r}^k ,..., \nu_{\mu_n,r}^k ],
	\end{equation}
	
	where $k$ is the threshold used to build the map.
	
	In order to capture information in different scales, a feature vector $\psi_r$ considering different values of thresholds $k$ is shown in Equation \ref{eq:vector3}. This vector consists of a concatenation of the vectors calculated using $ \varphi^k_r$, for different $k$ values: 
	
	\begin{equation}\label{eq:vector3}
	\psi_r= [\varphi^{k_0}_r, \varphi^{k_1}_r,..., \varphi^{k_m}_r].
	\end{equation}
	
	Finally, a feature vector $\upsilon$ can be considered for the two rules of movement. The final vector $\upsilon$ contains tourist information from different thresholds and rules of movement, as follows:
	
	\begin{equation}\label{eq:final}
	\upsilon = [\psi_{min},\psi_{max}].
	\end{equation}

	\section{Experiments and Results} \label{sec:exp}

	In this section, we present the experiments and results obtained by the proposed method. The experimental setup used is described in the following sections. We also present the influence of the method parameters for the classification task. Finally, the proposed method was compared with other methods of the literature on image database.

	\subsection{Experimental setup}\label{sec:expSetup}
	
	Experiments have been conducted on Brodatz database. This database was made using images from the book of Brodatz \cite{tuceryan1998texture},  a set of images widely used as benchmark for texture analysis. In this experiment, a total of 1110 images, divided into 110 classes with 10 samples each, were used. Each image has 200$\times$200 pixels with 256 gray scales.
	A feature vector was extracted for each image and a classification process using Linear Discriminant Analysis (LDA) \cite{everitt1993principal} was performed. LDA is a well-known method for estimating a linear subspace with good discriminative properties \cite{Backes200954,Backes20111684}.  LDA is also very used to evaluate method based on walks.
	A scheme 10-fold cross-validation was adopted for separation of the samples of test and training.
	This scheme splits the samples into 10 folds with the same number of sample per fold. One fold is used for test while the remaining folds are used for supervised training. This process is performed for all folds.
	
	

	\subsection{Parameter Evaluation}
	
\begin{figure}[!htbp]
		\centering
		\subfigure[Different memory sizes $\mu$]{\label{fig:memoryr} \includegraphics[width=0.5\textwidth]{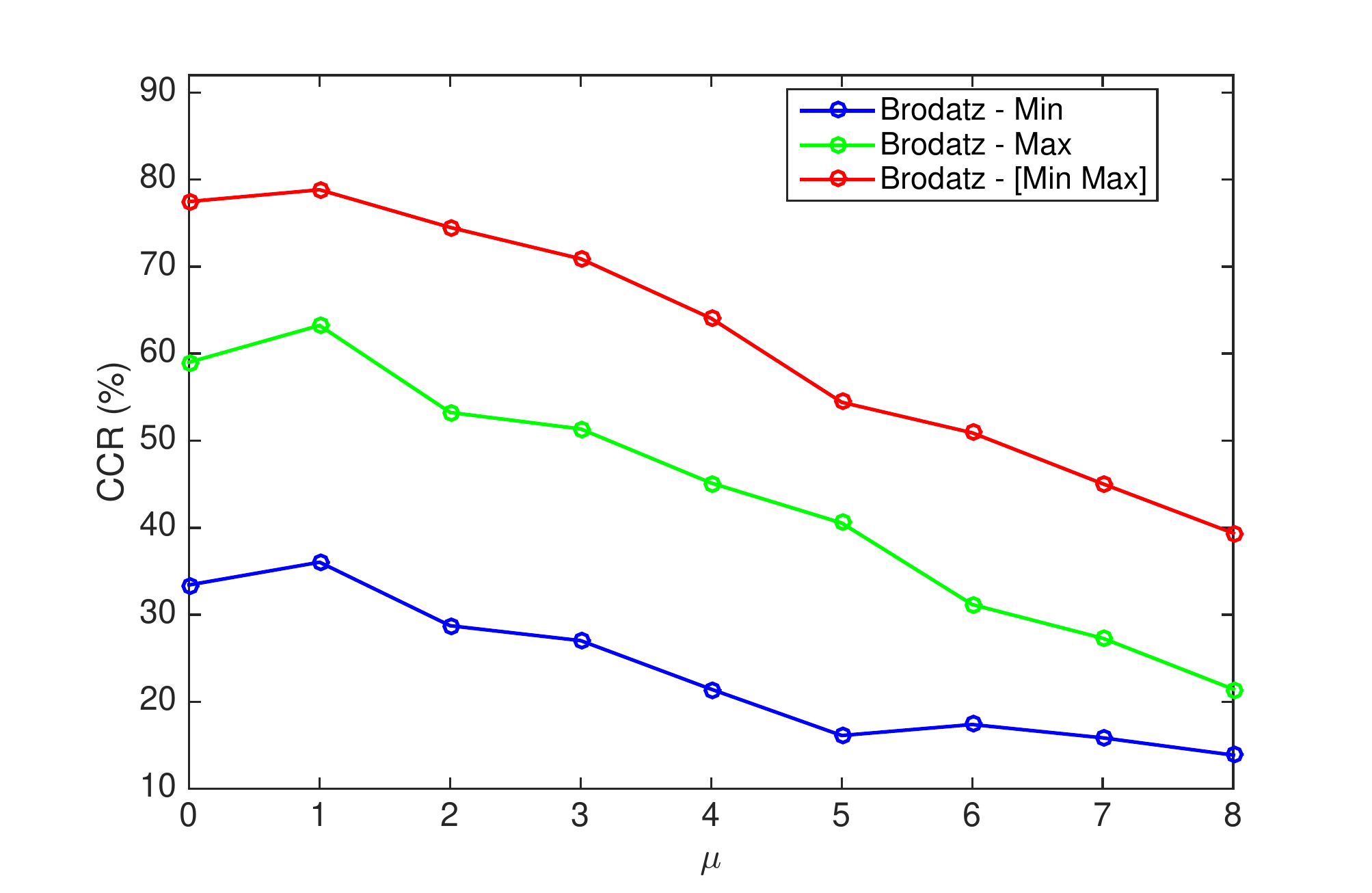}}\\
		\subfigure[Combining memory sizes $\mu$ ]{ \label{fig:memory_comb} \includegraphics[width=0.5\textwidth]{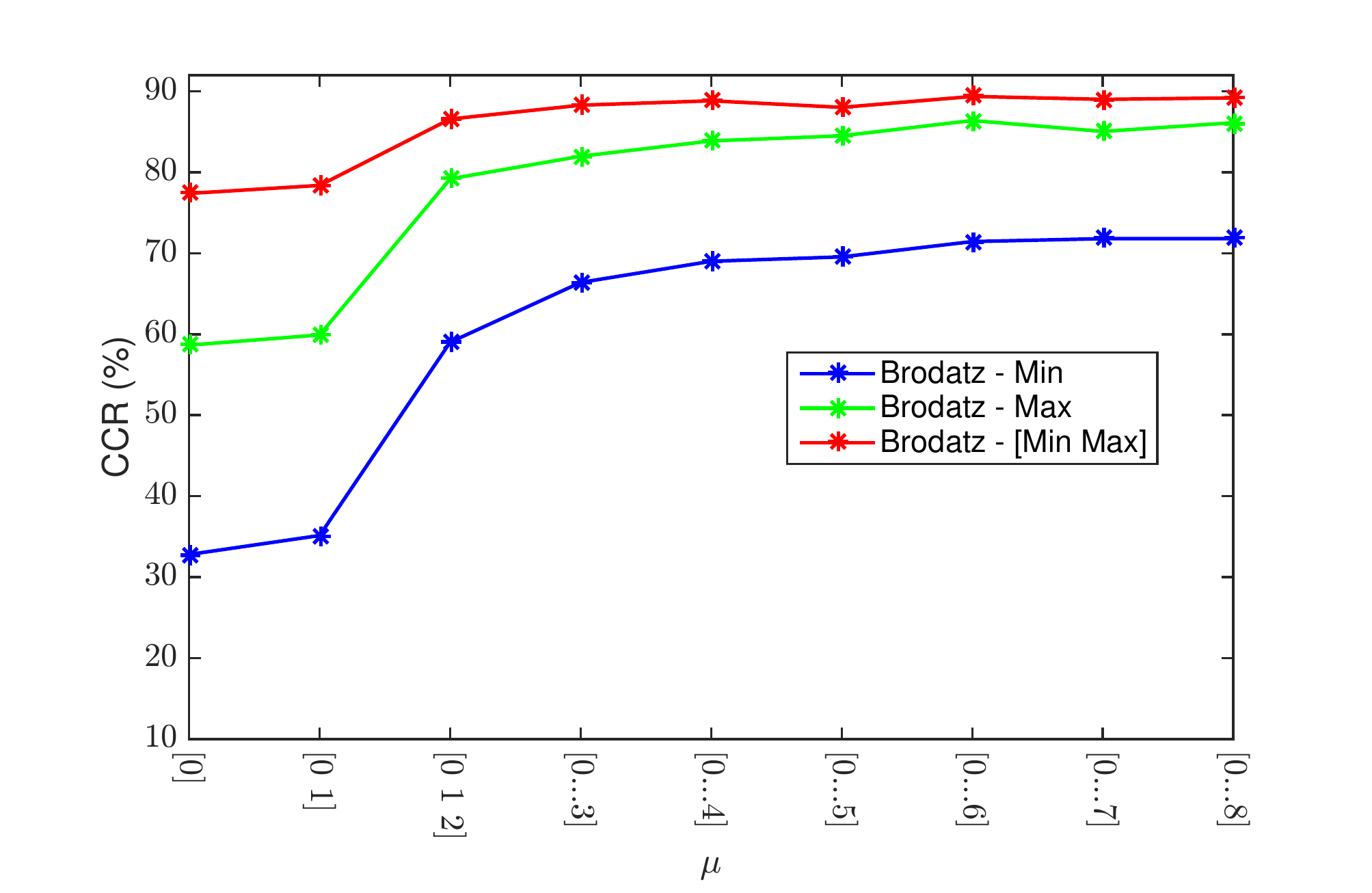}} 
		\caption{Correct classification rate as a function of different memory sizes $\mu$ and rule of movement $r$ .}
		\label{fig:ccr_u}
	\end{figure}

	In this section, we evaluate the influence of the parameters of the proposed method in texture recognition. The parameters are the (i) set of memory values $\mu$, (ii) rules of movement $r$, and (iii) set of thresholds $k$. At first, each feature vector was evaluated in order to determine the set of memory values that best characterizes the texture.
	This analysis was performed through the original map used by the traditional methods of DPSW, i.e., the threshold is in function of $k=0$.
	Figure \ref{fig:memoryr} shows the correct classification rate (CCR) for different memory values on Brodatz database. As we can see, small memory values provide best CCR compared with high memory values. Also, the rule of movement $max$ obtained a better CCR compared to the rule of movement $min$. Thus, attractors formed in heterogeneous regions have more importance in recognition task \cite{backes2010texture}.
	The characteristics of these regions are the high presence of contours or changes in texture patterns.
	However, the concatenation of the rules of movement $min$ and $max$ feature vectors achieved better results. This corroborates that heterogeneous and homogeneous image information is important in the recognition task. 
	
	Previous studies \cite{backes2010texture,Backes20111684} showed that the combination of memory values can achieve interesting results. The CCR for different combination of memory values is shown in Figure \ref{fig:memory_comb}. Notice that the best results are obtained for the concatenation of few memory values. 
	As the memory combination increases, the CCR tends to increase. This occurs due to two motives: more memory values lead to more descriptors in the feature vector and different memory values influence the way the walks are performed by the tourist \cite{backes2010texture}.
	As can be seen in the figure, the best CCR of 89.36 \% was obtained with the combination of memory values $\mu= [0, 1, 2, 3, 4, 5, 6]$. Therefore, the feature vector $\varphi^k_r= [\nu_{0,r}^k , \nu_{2,r}^k ,..., \nu_{6,r}^k ]$ is defined. 
	
\begin{figure}[!htbp]
		\centering
		\subfigure[Different threshold values $k$]{\label{fig:limiar} \includegraphics[width=0.5\textwidth]{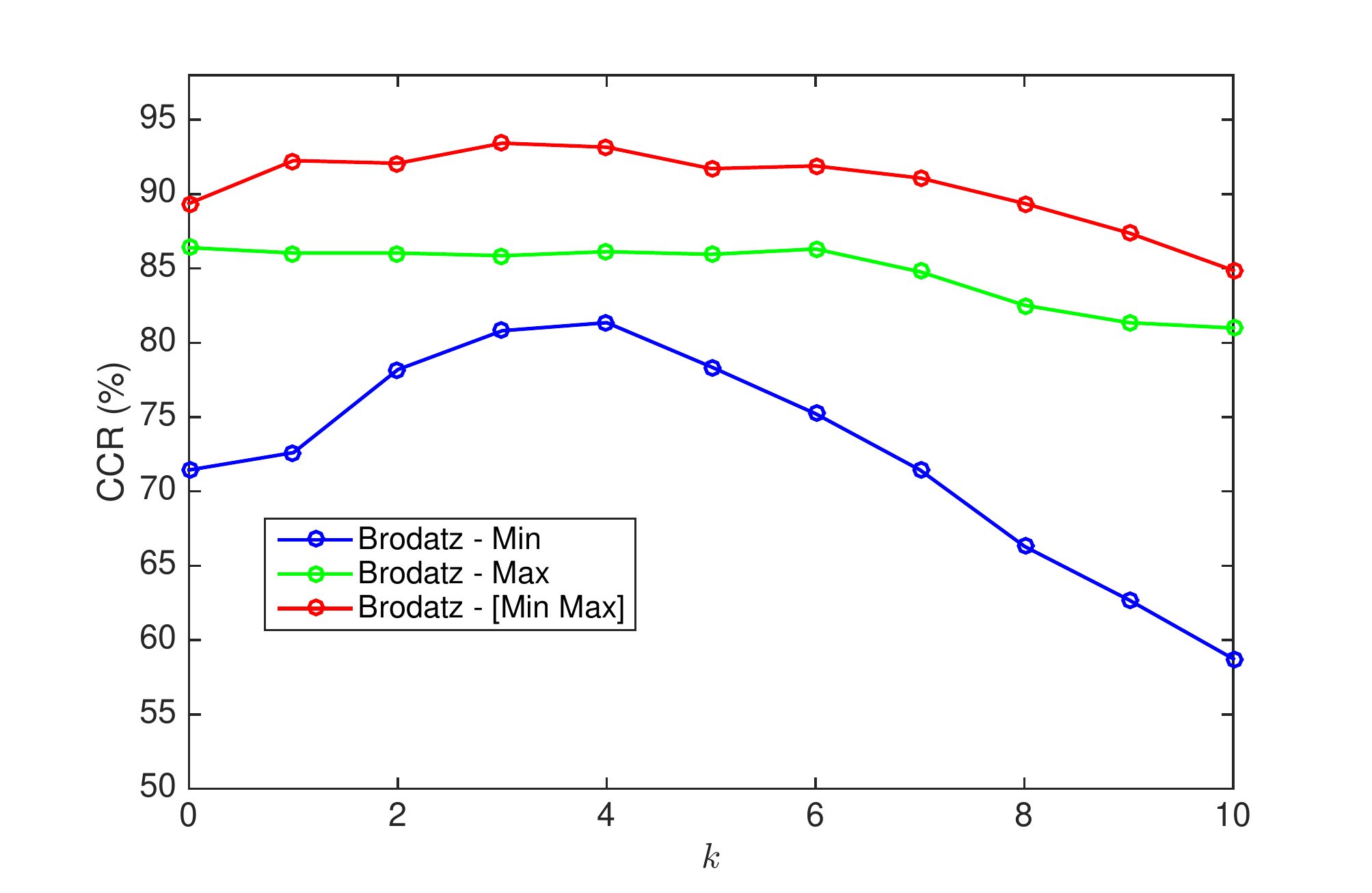}}
		\subfigure[Combining threshold values $k$]{ \label{fig:limiar_comb} \includegraphics[width=0.5\textwidth]{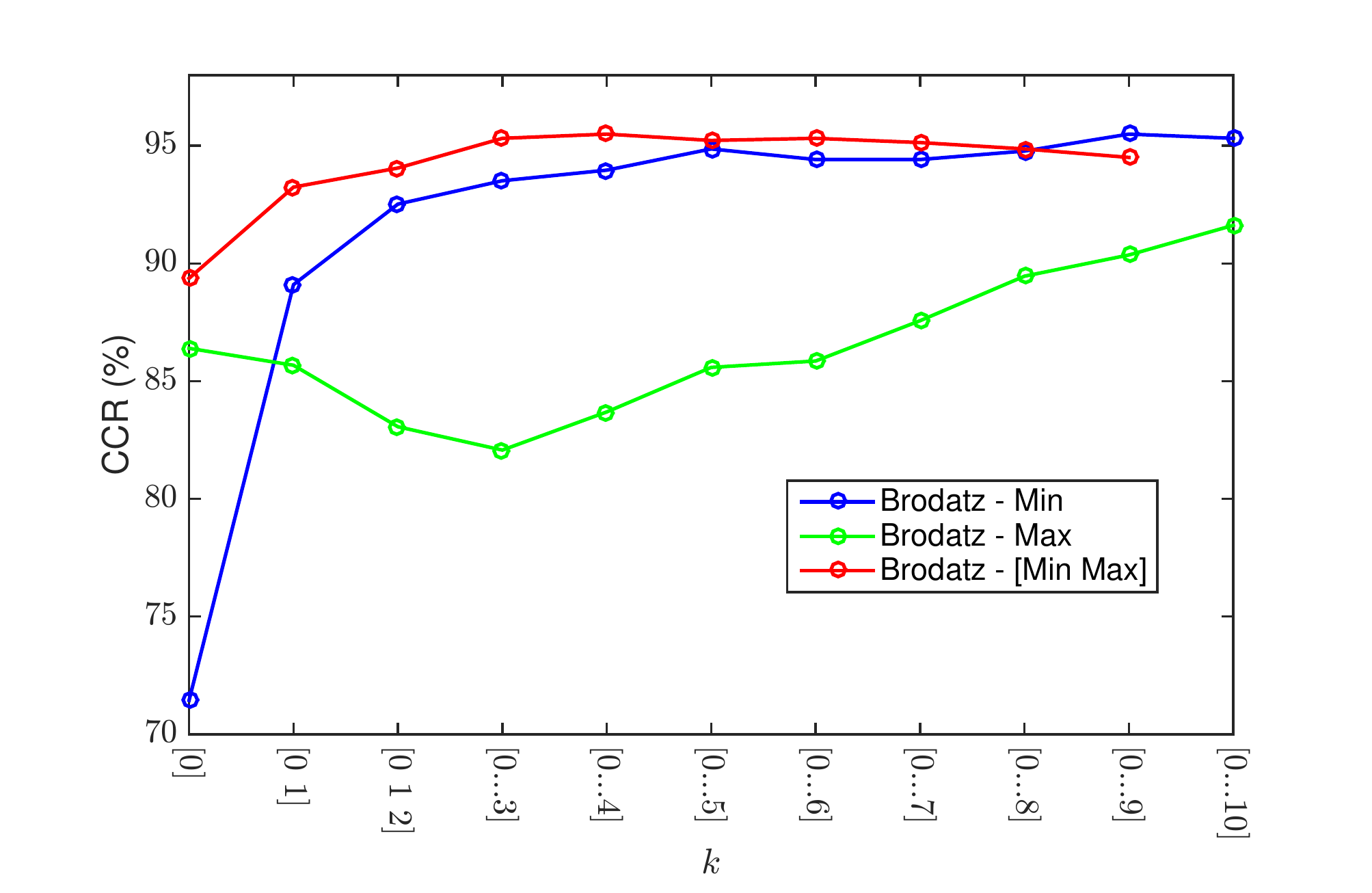}} 
		\caption{Correct classification rate as a function of different thresholds values $k$ and rule of movement $r$ .}
		\label{fig:limiares}
	\end{figure}

	From the set of memory values obtained in the analysis above, we evaluate the proposed method for different thresholds values and rules of movement. Here, we use an increment $i=10$ and $i=20$ for the thresholds $t_{min}$ and $t_{max}$, respectively. This increment value $i$ was defined in this paper based on experimental tests. Using this increment is possible to analyze different scales. Thus, the thresholds are defined as $t_{min} = k*10$ and $t_{max} = 255 -  k*20$. The threshold is evaluated in Figure \ref{fig:limiares} as a function of $k$. Note that for $k=0$ the threshold is equal to $t_{min}=0$ in the rule of movement $min$ and equal to $t_{max}=255$ in the rule of movement $max$. In this case, the map for the tourist to perform the walk will be the original. Figure \ref{fig:limiar} shows the CCR for different threshold values. Notice that, as the threshold is increased the CCR obtains an improvement for all rules of movement compared to traditional method ($k=0$). This shows that new information is extracted from the image with the use of thresholds on the neighborhood.  Using both rules of movement combined, the proposed method achieved the best result.
	
	For performing a multi-scale analysis, an interesting approach is to combine features extracted from different threshold values. The CCR for concatenation of thresholds is shown in Figure \ref{fig:limiar_comb}.
	As we can see, combining few thresholds obtains a significant improvement in the CCR compared to the original method.
	As the threshold increases, the neighborhood of the pixels is reduced, i.e., it is more difficult to find a set of pixels to compose an attractor. In so many cases, the tourist performs longer walks to find an attractor or reaches pixels without neighbors.
	Thus, the use of different threshold values provides a better exploration of image context, and it enables us to capture texture details in both micro- and macro-scales.
	This mechanism improves the capacity of discrimination of the proposed feature vector \cite{backes2010texture}. As can be seen in Figure \ref{fig:limiar_comb}, the best CCR of 95.5\% was obtained for the concatenation of the feature vector $\psi_{min}= [\varphi^0_{min}, \varphi^1_{min},\varphi^2_{min},...,\varphi^9_{min}]$.
	Once each feature vector $\varphi^k_r$ has size equal to 4, the vector $\varphi^k_r= [\nu_{0,r}^k , \nu_{1,r}^k ,..., \nu_{6,r}^k ]$ has size equal to 28. Thus, the final feature vector $\psi_{min}= [\varphi^0_{min}, \varphi^1_{min},\varphi^2_{min},...,\varphi^9_{min}]$ has size equal to 280.
	
	\subsection{Comparison with other methods}    
	
	In order to provide a better evaluation of the proposed method, we perform a comparison with other texture methods of the literature.
	In this comparison, it was used the best result achieved by our method.
	The same experimental setup described in Section \ref{sec:expSetup} was used for the experiments performed with literature methods.
	The methods used in this comparison are Fourier descriptors \cite{azencott1997texture},  co-occurrence matrices (GLCM) \cite{haralick}, Gabor filters \cite{Bovik199055}, local binary pattern rotate invariant and uniform (LBP-RIU) \cite{ojala2002multiresolution}, local oriented statistics information booster (LOSIB) \cite{losib}, texture analysis using graphs generated by DPSW (G-DPSW) \cite{Backes20111684}, segmentation-based fractal texture analysis (SFTA) \cite{sfta} and local binary pattern variance (LBPV) \cite{Guo2010706}. Besides that we compare our method with the traditional deterministic partially self-avoiding walk \cite{backes2010texture}.
	Table \ref{tab:comp} presents comparative results for the proposed method and traditional texture methods obtained on the Brodatz database. 
	To evaluate the methods, we use the CCR and standard deviation in parentheses.
	Experimental results indicate that the proposed method improves recognition performance from 89.36\% to 95.50\% over the traditional DPSW. The results indicate also that the proposed method outperformed other methods of the literature, except the GLCM method, whose the results were very close, 95.50\% ($\pm$ 1.47) for the proposed method and 94.59\% ($\pm$ 1.58) for the GLCM method.

	\begin{table}[]
		\centering
		\caption{Comparison results for literature methods in the Brodatz database.}
		\label{tab:comp}
		\begin{tabular}{|c|c|}\hline
			\textbf{Methods} & \textbf{ CCR (\%) } \\ \hline
			Fourier & 87.56 ($\pm$ 2.39)   \\ \hline
			GLCM & 94.59 ($\pm$ 1.58)    \\ \hline
			Gabor& 91.44 ($\pm$ 2.44)    \\ \hline
			LBPV & 88.10 ($\pm$ 2.74)   \\ \hline
			LBP RIU & 89.09 ($\pm$ 4.01)   \\ \hline
			LOSIB & 70.45 ($\pm$ 3.47)   \\ \hline
			SFTA & 83.87 ($\pm$ 3.53)   \\ \hline
			G-DPWS & 90.35 ($\pm$ 3.32)   \\ \hline
			DPSW traditional & 89.36 ($\pm$ 2.39)   \\ \hline        
			\textbf{Proposed method} & \textbf{95.50 ($\pm$ 1.47)}   \\ \hline
		\end{tabular}
	\end{table}
	
	\section{Conclusions}
	\label{sec:conc}

	In this paper, we have proposed an effective texture analysis method based on deterministic partially self-avoiding walks and thresholds. In this method, tourist walks are performed in maps built with different thresholds over the neighborhood of the pixels. For each image, the features using various thresholds values are concatenated to compose a feature vector to texture classification. We showed promising results obtained on well-known public image database. In the Brodatz database, experimental results indicate that the proposed method improves recognition performance from 89.36\% to 95.50\% over the traditional DPSW method. The proposed method also provides excellent results compared to other traditional texture methods.
	
	\section*{Acknowledgment}
	
	The authors gratefully acknowledge support from Coordination for the Improvement of Higher Education Personnel - CAPES (PROEX-9254772/M), CNPq (Grant Nos. 307797/2014-7 and 484312/2013-8), FAPESP (Grant No. 14/08026-1) and Fundect (No. 071/2015).

\bibliographystyle{model1-num-names} 


\end{document}